# Driver Action Prediction Using Deep (Bidirectional) Recurrent Neural Network

Oluwatobi Olabiyi, Eric Martinson, Vijay Chintalapudi, Rui Guo
Intelligent Computing Division
Toyota InfoTechnology Center USA
Mountain View, California, USA
{oolabiyi, emartinson, achintal, rguo}@us.toyota-itc.com

*Abstract*—Advanced driver assistance systems (ADAS) can be significantly improved with effective driver action prediction (DAP). Predicting driver actions early and accurately can help mitigate the effects of potentially unsafe driving behaviors and avoid possible accidents. In this paper, we formulate driver action prediction as a timeseries anomaly prediction problem. While the anomaly (driver actions of interest) detection might be trivial in this context, finding patterns that consistently precede an anomaly requires searching for or extracting features across multi-modal sensory inputs. We present such a driver action prediction system, including a real-time data acquisition, processing and learning framework for predicting future or impending driver action. The proposed system incorporates camera-based knowledge of the driving environment and the driver themselves, in addition to traditional vehicle dynamics. It then uses a deep bidirectional recurrent neural network (DBRNN) to learn the correlation between sensory inputs and impending driver behavior achieving accurate and high horizon action prediction. The proposed system performs better than other existing systems on driver action prediction tasks and can accurately predict key driver actions including acceleration, braking, lane change and turning at durations of 5sec before the action is executed by the driver.

## I. Introduction

Driving is an indispensable component of daily life. Over decades, tremendous effort has been dedicated to improving the safety and efficiency of driving. Although earlier advanced driver assistance systems (ADAS) primarily focused on sensing dangerous external environmental factors that might impact safe driving, recent efforts are now shifting to incorporating driver intention into the system [1-3] (e.g. is the driver already planning on braking in the near future?). Several of these recent studies have demonstrated that predictive driving assistant systems that are aware of the driving patterns of intentional driver behavior will be very helpful by providing early notification to further mitigate dangerous driving maneuvers [1-11].

While the high observability of the interaction between driver and vehicle makes it easy to recognize driver action when taking place, predicting the action before it takes place is still a daunting task as it involves intricate, multi-dimensional dynamics[1].

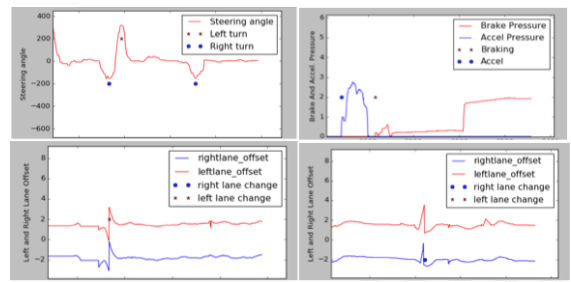

Fig. 1. An example of timeseries data that capture common changes (anomaly) in driver behavior. **Upper Left:** Left and right turn; **Upper Right:** Braking and Acceleration; **Lower Left**: Left lane change; **Lower Right**: Right lane change.

In this paper, we formulate the driver action prediction as a timeseries anomaly prediction problem. Therefore, our proposed framework contains two main components. First is the anomaly detection or driver action recognition system. Fig. 1 shows the detection of common driver actions related to pedalling and steering as timeseries anomalies. The second component is the anomaly prediction or driver action prediction (DAP) system which involves predicting earlier detected anomalies before they occur again in the future. Our prediction system uses deep Bidirectional Recurrent Neural Network (DBRNN) consisting of multiple Long-Short Term Memory (LSTM) units and/or Gated Recurrent Units (GRU) cells that learns to identify as early as possible, the spatial-temporal dependencies in timeseries data with respect to future driver action. Therefore, when those dependencies occur again in the data stream, our model is able to accurately predict an impending driver action before the driver takes the action. Our choice of prediction algorithm in conjunction with high observability of driver action enables the possibility of training the driver action prediction model both offline and online using the data obtained from the action recognition system.

The rest of the paper is organized as follows: in Section II we describe the related driver action prediction work, and Section III contains the system description while Section IV contains the deep RNN principle and structure. Driver-vehicle-environment (DVE) feature formulation is detailed in Section V, and experimental analysis on the real-world driving dataset is in Section VI. We summarize the entire work in Section VII.

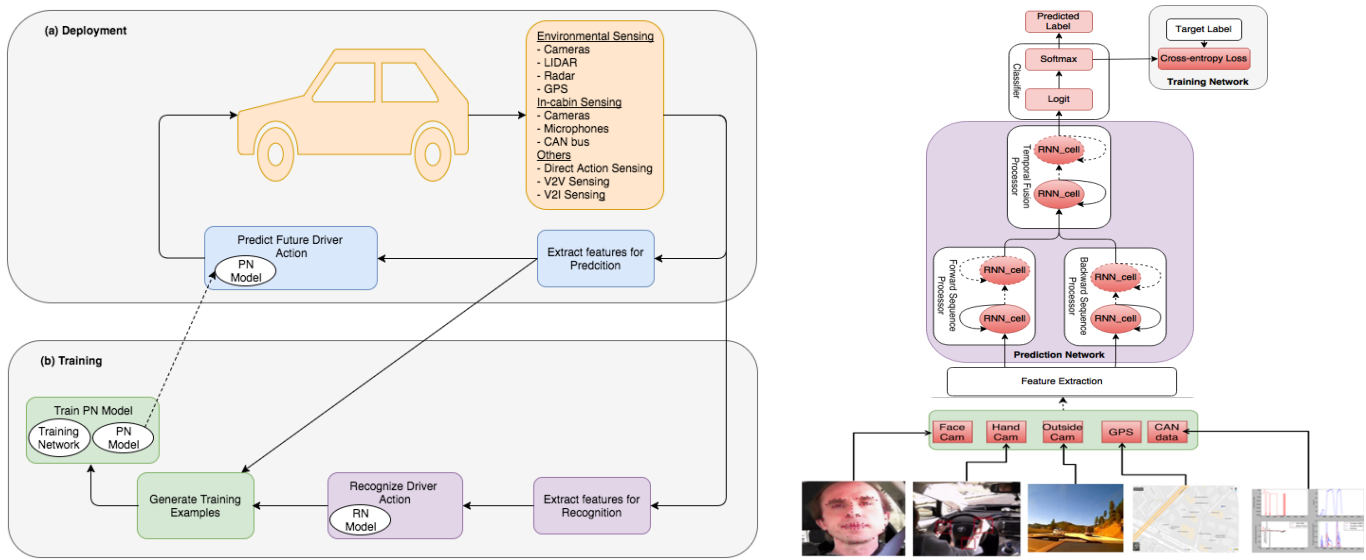

Fig. 2. **Left:** The Proposed DAP system. **Right:** The Proposed Prediction Network based on DBRN

## II. RELATED WORK

Most existing work on driver action prediction focuses on feature engineering in addition to basic temporal sequence modeling. For each action of interest, features that are temporally correlated to an action of interest are selected based on prior or expert knowledge. For example, in [7], vehicle speed measurements from the CAN-bus, along with the traffic light sensing data, are adopted for predicting driver braking behavior with a 3sec time window. In [8], yaw and steering-wheel angles were considered for lane departure analysis. Vehicle velocity, steering angle, and external GPS information were used for modelling the lane-change maneuver in [9]. Similarly, vehicle dynamics are used for identifying driver intention near intersections as proposed in [10]. The long-term trajectories are also informative for driver intention analysis. In [11], a Probabilistic Finite State Machine was used to model driver intention from the trajectory data. Others also studied the use of extracted features from vision-based sensors such as cameras. In [12,13], a front-facing camera was used to predict different driving behaviors. The authors also used ego-vehicle dynamics and lane information extracted from cameras to recognize driving related events.

Driver behavior modeling can also involve advanced control and machine learning techniques. In some of the existing literature [14-17], driving models were built using optimal control theory for analyzing steering behavior, tracking the driving path or estimating the route. In [18-19], Bayesian reasoning was used to recognize driver intention with complex input feature dimension. Other high performance models [20-22] used dynamic Bayesian networks, or Hidden Markov Model (HMM) and its modified structures to achieve remarkable achievements in various applications.

However, formulating DAP as a timeseries anomaly prediction problem reveals that hand engineered features might not yield the best result. This is because the DAP system should be capable of extracting relevant sensing information that is most useful in predicting a future driver action. Therefore, with sensor-rich input, there's need for a learning framework that can take advantage of this. Although the authors in [2] explore sensor-rich input for lane change prediction, the learning framework based on relevant vector machine (RVM) has limited feature extraction capability. The authors in [3] addressed this problem by using LSTM based recurrent neural network (capable of extracting richer discriminative features) with spatial sensory fusion to improve performance. They also include turn prediction in their work.

In comparison, this work explores a richer sensory input than in [2] and [3]. Our model can also predict both steering and braking/acceleration actions unlike [2] and [3] that focused only on steering actions. More importantly, using a windowing approach similar to [2], we employ deep bidirectional RNN to extract richer and more stable representation of the sensory input. This is the core algorithmic difference between this paper and the proposition in [3]. While the double LSTM units used in [3] achieved spatial fusion by assigning in-vehicle features to one LSTM and outside vehicle features to the other, ours assign all feature to both LSTM units but one unit processes the sensory sequence from past to current and the other from future to current. This fusion of different temporal directions enables the model to find causes of future action more accurately and much earlier than achievable in [3].

## III. SYSTEM OVERVIEW

The purpose of the proposed DAP system is to correctly estimate driver intent as early as possible before the driver takes action. The system shown in Fig. 2 is capable of modeling the prediction of any recognizable driver action. The proposed DAP system depicted on the left of Fig. 2 consists of sensing module, feature extraction modules, action recognition module, action prediction module and training module.

TABLE I. EXAMPLE OF SPLIT OF SENSORY FEATURES BETWEEN ACTION RECOGNITION AND PREDICTION SYSTEMS

| Modality | Recognition | Prediction |
|---|---|---|
| CAN Bus Data | Brake and accelerator pressure, steering angle | Brake and accelerator pressure, gear positions, steering angle, velocity and acceleration, engine rpm, elevation (8) |
| Face Camera | None | Head pose (9) |
| Hand Camera | None | Hand positions, movement status and directions (19) |
| Dash Camera | Left lane offset<br>Right lane offset | adjacent lane availability, relative position and velocity of road objects to driver's and adjacent lanes, left and right lane offset and curvature (12) |
| GPS + Map | None | Distance and direction from nearest intersection (2) |

The sensing module handles sensing. The feature extraction modules independently extract features which can be used to recognize and predict driver actions as shown in Table 1. The driver action prediction module continuously predicts future driver action by running the prediction network (PN) model on the features extracted for prediction. On the other hand, the driver action recognition module runs the recognition network (RN) model on the features extracted for recognition in order to recognize driver actions. Lastly, the training module is retroactively combining features extracted for prediction and the recognized action label to generate training examples, which are then used to train the PN model. Our system uses off-the-shelf feature extraction systems and therefore our main algorithmic contribution is the prediction network shown on the right of Fig. 2.

IV. DBRNN FOR DRIVER ACTION PREDICTION

The key component in this work is the use of DBRNN framework shown on the right hand side of Fig 2 to model future driver action. The RNN cell can be simple, long-short term memory or gated recurrent unit RNNs. Since our goal is to predict driver action before it takes place and assuming that both negative and positive states can be recognized correctly by the recognition network, our DBRNN framework focuses on modeling the transition between negative state (none-action state) and positive state (action state) in feature sequences.

Consider a stream of driving data, we use the recognition system to mark the beginning of each driver action. Suppose an action event occurs at time $t_a$. We define an action event random variable, $a \in \{a_1,..,a_K\}$ and a timing random variable $\tau \in \{\tau_1,..,\tau_L\}$. Note that $a_K$ is the normal (negative) action state while the rest are anomalous. We define two parameters; $d$, the maximum horizon (transition period) and $T$, the maximum sequence length per prediction. At any time $t$, the prediction system models the joint conditional probability

$$y_t = P(a = a_k, \tau = \tau_l | X_t; \Theta) \quad (1)$$

where $X_t$ is the model input time $t$, and $\Theta$ is the model parameters. In [2], $X_t = (x_{t-T-1}, ..., x_t)$ where $x_t$ is the input sample vector at time $t$, $\tau \in (t < t_a - d, t = t_a - d)$, $d =$ 2.5sec and $T$ = 2sec. However, to overcome the timing problem in [2] (i.e. difficulty in precise timing of causes of driver action), [3] using RNN model relaxed the timing constraint and substitute $\tau \in (t < t_a - d, t_a - d \leq t < t_a)$, $X_t = x_t$, $T$ = 0.8sec (1 sample) and $d$ = 4.8sec (7samples). Please note that the internal state of the RNN is used in [3] to provide context from one timestamp to the other.

In order to limit arbitrary context propagation and therefore improve performance over [3], our DBRNN network uses a window similar to [2] to model the conditional probability in (1) with $\tau \in (t < t_a - d, t_a - d \leq t < t_a)$ and $X_t = (x_{t-T-1}, ..., x_t)$ and $d = T = 5sec$ (50samples) in all our evaluations.

A. Network Architecture

The prediction network is implemented with a Recurrent Neural Network (RNN). The neural network takes the temporal sequence of observation $X_t = (x_{t-T-1}, x_{t-T-2}, ..., x_t)$ as input, and generates a sequence of vectors $(h_{t-T-1}, x_{t-T-2}, ..., h_t)$ via a non-linear activation. The output vectors are hidden variables that describe the distribution of the observation. Different from the traditional feed-forward neural network, a RNN also considers historical information or so-called "memory". The label is determined by a non-linear mapping over these hidden variables. The mathematical formulations are described in the following equations.

$$h_t = H(W_{xh}x_t + W_{hh}h_{t-1} + b_h) \quad (2)$$

$$y_t = F(W_{hy}h_t + b_y) \quad (3)$$

where $H$ is a non-linear function which is chosen from *sigmoid* and *tanh* conventionally. $F$ is usually chosen from *sigmoid* or *softmax* if the output label are just independent or mutually exclusive respectively. $W$ and $b$ are associated weight (e.g. $W_{xh}$ is the input-hidden weight matrix), and bias (e.g. $b_h$ is hidden bias vector) matrices respectively. These parameters are learned from the training data.

Recently, researchers demonstrated a problem with simple RNN structures where the gradient vanishes in training. To solve this problem, the most popular replacements are Long Short-Term Memory (LSTM) and Gated Recurrent Unit (GRU) cells that can maintain their state over time, resulting in captured long term dependencies within time series data [23-27]. The versions of LSTM and GRU cell used in this paper are respectively given in [3, 25] and [26]. We use the following shorthand notation to respectively denote the LSTM and GRU cell operations.

$$h_t = H_{LSTM}(x_t, h_{t-1}, c_{t-1}) \quad (4)$$

$$h_t = H_{GRU}(x_t, h_{t-1}) \quad (5)$$

The main difference between LSTM and GRU is that the content of GRU cell memory is always exposed to the output and therefore it is easier to implement since it requires fewer network parameters. In summary, our RNN cell $H_\chi(.)$ can be LSTM (Eq. 4) or GRU cell (Eq. 5) RNN.

## B. Bidirectional RNN

One shortcoming of conventional RNNs is that they are only able to make use of previous context. However, with a windowed approach where whole temporal context is available, there is no reason not to exploit future context as well. As shown in right hand side of Fig. 2, the Bidirectional RNNs (BRNNs) using basic RNN units do this by processing the data in both directions with two separate hidden layers, which are then fed forward to the same output layer. A BRNN computes the forward hidden sequence $\vec{h}_t$, the backward hidden sequence $\overleftarrow{h}_t$, and the output sequence $y_t$ by iterating the backward layer from $t-T+1$ to $t$, the forward layer from $t$ to $t-T+1$ and then updating the output layer:

$$\vec{h}_t = H(W_{x\vec{h}}x_t + W_{\vec{h}\vec{h}}\vec{h}_{t-1} + b_{\vec{h}}) \quad (6)$$

$$\overleftarrow{h}_t = H(W_{x\overleftarrow{h}}x_t + W_{\overleftarrow{h}\overleftarrow{h}}\overleftarrow{h}_{t-1} + b_{\overleftarrow{h}}) \quad (7)$$

$$y_t = F(W_{\vec{h}y}\vec{h}_t + W_{\overleftarrow{h}y}\overleftarrow{h}_t + b_y) \quad (8)$$

Facilitating LSTM with this bidirectional learning capability enables the system to access long-range context in both input directions and thus enrich sequential data understanding.

## C. Deep RNN

With the potential to disentangle complicated temporal dependencies, deep RNNs can be created by stacking multiple RNN hidden layers on top of each other, with the output sequence of one layer forming the input sequence for the next, as shown in the LHS of Fig. 2. Assuming the same hidden layer function is used for all $N$ layers in the stack, the hidden vector sequences $h^n$ are iteratively computed from $n = 1$ to $N$ and $t - T+1$ to $t$:

$$h_t = H(W_{h^{n-1}h^n}h_t^{n-1} + W_{h^nh^n}h_{t-1}^n + b_{h^n}) \quad (9)$$

$$y_t = F(W_{h^Ny}h_t^N + b_y) \quad (10)$$

where $h^0 = x$.

## D. Deep Bidirectional RNN

Deep bidirectional RNNs can be implemented by replacing each hidden sequence $h^m$ with the forward and backward sequences $\vec{h}^m$ and $\overleftarrow{h}^m$, and ensuring that every hidden layer receives input from both the forward and backward layers at the level below. In the proposed network on the right of Fig. 2, a deeper architecture is obtained by replacing each RNN unit in the deep bidirectional RNN by a deep unidirectional RNN. Hence, for $M$ stacked BRNN and each of forward and backward cell having $N$ stacked RNN configuration, we obtain a 2x$M$x$N$ RNN unit system. However, we found out that very deep configurations do not yield better performance on our limited dataset as they only over-fit the training data. Therefore, for the dataset considered here, we only use single bidirectional LSTM stacked with single unidirectional GRU cell yielding a three RNN unit system, i.e.

$$(\vec{h}_t^1, \vec{c}_t^1) = H_{LSTM}(x_t, \vec{h}_{t-1}^1, \vec{c}_{t-1}^1) \quad (11)$$

$$(\overleftarrow{h}_t^1, \overleftarrow{c}_t^1) = H_{LSTM}(x_t, \overleftarrow{h}_{t-1}^1, \overleftarrow{c}_{t-1}^1) \quad (12)$$

$$h_t^2 = H_{GRU}([\vec{h}_t^1 : \overleftarrow{h}_t^1], h_{t-1}^2) \quad (13)$$

$$y_t = F(W_{h^2y}h_t^2 + b_y) \quad (14)$$

We however believe that larger dataset will benefit from deeper configuration.

## E. Sensory Fusion

Although our system uses multi-modal inputs we do not need to feed each input modality to its own RNN since the modalities are already pre-processed. Therefore, we only concatenated the features together, and pass it through the network as a single vector. We also evaluated the spatial sensory fusion demonstrated in [3], but it did not yield an improved performance with our deeper structure and bidirectional RNN.

## F. Network Training

We train the network with back propagation through time, using Tensorflow on an Nvidia GTX GPU running Ubuntu 14.04. The network weights and biases are adapted using the Adam optimizer with a learning rate of 1e-2 and decay of 0.1 per 100 epochs. The maximum epoch is set to 1000. The gradient on each RNN cell is clipped to max value of 10 to avoid gradient explosion problem. Each RNN cell has 64 hidden units.

## V. FEATURE EXTRACTION

The performance of machine learning methods is heavily dependent on the choice of data representation (or features) on which they are applied. For the data used in this paper, the sensory input of the DAP system is the multi-modality sequences derived from the following: (i). CAN-bus: steering angle, pedal pressures, yaw rate, speed, acceleration, road slope and engine rpm; (ii). Face camera: Head pose and eye gaze; (iii) Hand camera: hands positions, hands moving and on/off streering wheel status; (iv) Dash camera: adjacent lane availability, right and left lane offset, relative position and speed of road objects to the driver and adjacent lanes (v) GPS+map: distance and direction from nearest intersection. The way we have separated the incoming sensory data for recognition and prediction purposes is depicted in Table 1. The following is the detailed description of the features extracted for driver action prediction.

### A. Vehicle Dynamic Information

Monitoring the CAN bus is the most straightforward method of determining the vehicle's working status and the driver-vehicle interaction. Data are obtained by decoding the CAN codes according to the OBD2 protocols. We obtain an 8-dimension feature vector from the CAN-bus.

## B. Driver Behavior Information

### 1) Face Camera Feature Extraction

The driver's head motion is determined by analyzing video of the driver's face. The processing pipeline consists face detection, facial landmark tracking and feature extraction. Similar to [3], face detection and landmark tracking tasks are implemented with Constrained Local Neural Field (CLNF) model [28]. There are totally 68 landmarks are extracted from the detected face. Based on these landmark tracks, the driver's face movements and rotations are represented with histogram features. In particular, the matching landmarks between successive frames and their per-pixel horizontal changes are calculated to model the movements and rotations. Finally, the mean movement and histograms of horizontal and angular motions with bins $[\leqslant -2, -2$ to $0, 0$ to $2, \geqslant 2]$ and $[0$ to $\pi/2, \pi/2$ to $\pi, \pi$ to $3/2\,\pi, 3/2\,\pi$ to $2\pi]$ are calculated. The final result is a 9-dimensional feature vector per sample to represent face features.

### 2) Hand Camera Feature Extraction

Hand gesture is another important driving behavior descriptor. The location and movement of the hands on the steering can be highly correlated to impending lane change or turn and in some cases even braking and acceleration. We adopt a FRCNN facilitated hand detection algorithm [29] and extracted the following features: hand positions, distance, relative angle towards the steering center, relative position on the steer wheel, motion, moving distance and moving direction. Overall, 19-dimensions per sample encode hand information during driving.

## C. Driving Environment Information

The environment provides the most influential driving context, affecting driver's judgment and driving maneuver change. Features describing the environment were constructed from the outputs of a Mobileye system [30] processing dash camera images. The extracted features include: (1) two binary features indicating whether a lane exists on the left side and on the right side of the vehicle and their respective offset and curvature; (2) object (pedestrian or vehicle) position and speed information in driver and both adjacent lanes. An 12-dimension feature vector is obtained from the dash camera.

We also compare the vehicle GPS coordinates with street maps to build a 2-dimensional representation of proximity to an intersection. This consists of: (1) a 3-factor feature indicating if the vehicle is at (within 40 meters), approaching or departing a road artifact such as intersections, turns, highway exists, etc; and (2) the actual distance from the nearest intersection.

TABLE II. NUMBER OF POSITIVE AND NEGATIVE EXAMPLES (OBTAINED FROM THE RECOGNITION SYSTEM) USED FOR TRAINING(70%), CROSS-VALIDATION(15%) AND TESTING (15%)

| Driver Action | Positives | Negatives (Balanced) |
|---|---|---|
| Braking | 1033 | 1836 (1550) |
| Lane Change | Left-109, Right-125 | 4509 (188) |
| Turns | Left-264, Right-269 | 3516 (404) |

## VI. EXPERIMENTS AND RESULTS

### A. Data Collection

We collected about 35hr of driving data under different driving conditions across 5 drivers recorded in south San Fransisco bay area, California. The data collection vehicle is outfitted with three cameras, GPS and CAN bus data logger. The sampling rates across the sensors are 28-30fps, 1Hz and 80Hz for cameras, GPS and CAN bus respectively. For action prediction, we resampled the incoming data to 10Hz. The missing data are either extrapolated for float-value features or repeated with the nearest past value for factor-value features. For each of postive and negative samples a 5sec sequence is used yileding an input size of 50x50 per example ($T$=50).

To demonstrate the proposed system capabilities, we evaluate for braking, lane change and turn anomaly action prediction. Table 2 contains the detected number of anomalous actions and normal states using our proposed recognition system (before and after class balancing). In all cases, we compare the performance of the proposed DBRNN to unidirectional RNN system in [3] using a balanced dataset. Finally, we investigate the performance when training with data from an individual driver and compared to generic model of all five drivers.

### B. Braking Action Prediction

Fig. 3 plots performance for predicting braking events up to 5sec before braking is detected. The proposed DBRNN (Bi-LSTM Win) network gives comparable result to a unidirectional LSTM (Uni-LSTM) with a slight performance improvement at the beginning of the test sequence yielding a better prediction horizon. The unidirectional LSTM is obtained by removing the backward LSTM in Eq. (12) from the network. Overall, the proposed DBRNN system is able to achieve ~80% average accuracy, 70% true positive rate and 12% false positive rate 3sec from the braking event.

### C. Lane Change Action Prediction

In Fig. 4, we depict the performance of lane change event prediction. Here, the proposed Bi-LSTM network shows a significant performance improvement over the unidirectional

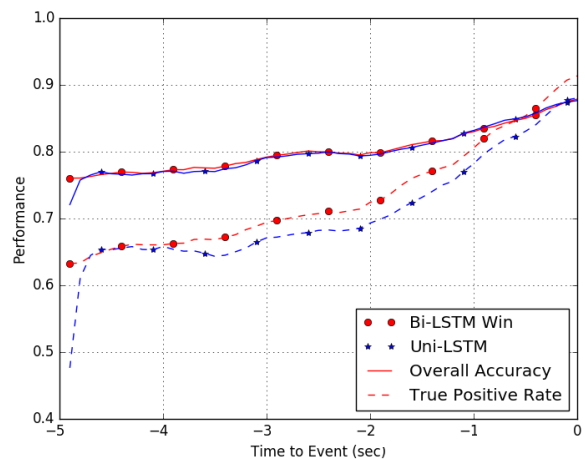

Fig. 3. Piecewise performance vs. time-to-event for braking predictions.

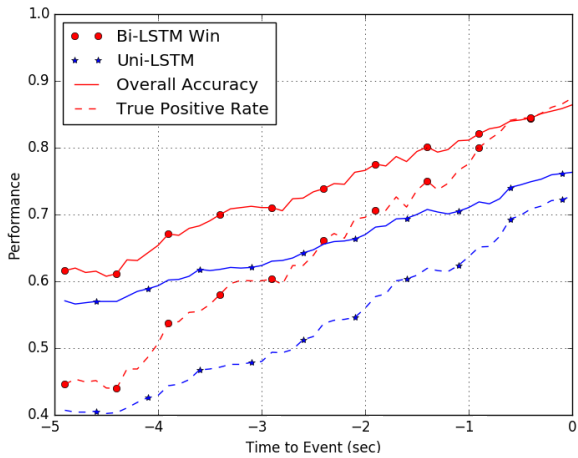

Fig. 4. Piecewice performace vs. time-to-event for lane change.

LSTM over the entire 5sec preceding the event. However, due to the fewer number of positive lane change actions, the overall performance is less than that of braking prediction. Please note that the true positive rate here is the average of positive rates for left and right lane change.

*D. Turning Action Prediction*

When predicting left and right turns, the proposed DBRNN system again shows better performance than the unidirectional version (Fig. 5). Both models perform better and worse compared to lane change and braking prediction respectively. It is also worth to note that the performance gap between DBRNN and unidirectional version is higher and lower compared to braking and lane change prediction respectively.

*E. Individual Driver Action Prediction*

Finally, we apply the predictive models to the braking data of an individual driver as opposed to all 5 drivers data used for the evaluations above. The overall accuracy comparison is depicted in Fig. 6. The result shows that both models perform better on an individual driver data compared to combined driver in Fig. 3. However, the proposed DBRNN performance improvement over the unidirectional version is higher with an individual driver.

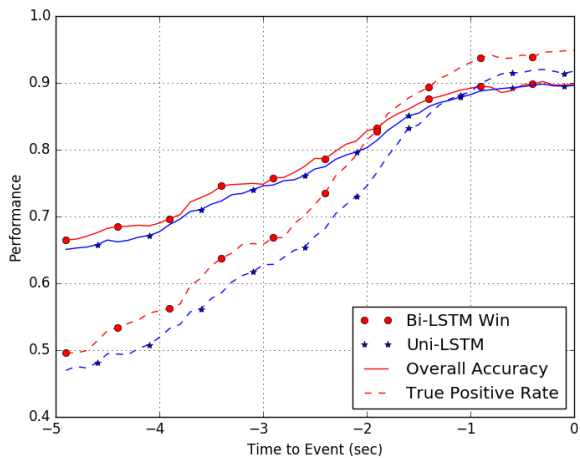

Fig. 5. Piecewice performace vs. time-to-event for turning.

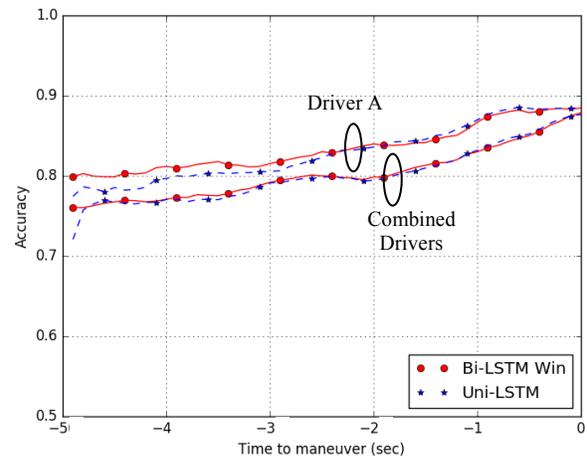

Fig. 6. Piecewice Overall Accurcy vs. time to event for braking prediction for individual and combined 5 drivers.

*F. Summary of Results*

In summary, the results indicate that the proposed DBRNN performs better than the unidirectional version across all but one evaluated scenarios. The exception is with predicting braking using a generic (vs individualized) model, when the two systems demonstrated comparable performance.

In general, the performance improvement appears to increase with reduced number of detected positive examples as well as on individual driver data. Also, the performance of both models increases with individual driver data compared to generic driver data. This highlights that our predictive model is able to learn with data from a single individual, making real-time training in the vehicle a real possibility for future predictive systems.

## VII. CONCLUSION

In this paper, we have proposed a novel DAP system that integrates both recognition and prediction systems. The proposed prediction system is based on DBRNN, which enables temporal fusion of both past and future context to learn the correlation between sensor data and future driver action. The proposed DBRNN system performs better than the state-of-the-art DAP system based on a unidirectional RNN, and is potentially trainable in vehicle for modeling individual drivers.

The high prediction accuracy and prediction horizon performance will enable new driver assistance capabilities to effectively alert drivers before taking dangerous driving actions. Our proposed prediction system can also be applied to a general class of anomaly prediction system. Our future works include improving system performance by extending input sensing modalities as well as extending system capability to directly predict both the anomalous event and time-to-event simultaneously.

ACKNOWLEDGMENT

The authors would like to thank Toyota Motor Corporation for sponsoring the research project.